# Sparse Distributed Memory using Spiking Neural Networks on Nengo

Rohan Deepak Ajwani
*BITS Pilani*
Goa, India
ORCID:0000-0003-3867-6172

Arshika Lalan
*BITS Pilani*
Goa, India
ORCID:0000-0002-8176-2431

Basabdatta Sen Bhattacharya
*BITS Pilani*
Goa, India
basabdattab@goa.bits-pilani.ac.in

Joy Bose
*Ericsson*
Bangalore, India
joy.bose@ericsson.com

*Abstract*— We present a Spiking Neural Network (SNN) based Sparse Distributed Memory (SDM) implemented on the Nengo framework. We have based our work on previous work by Furber et al, 2004, implementing SDM using N-of-M codes. As an integral part of the SDM design, we have implemented Correlation Matrix Memory (CMM) using SNN on Nengo. Our SNN implementation uses Leaky Integrate and Fire (LIF) spiking neuron models on Nengo. Our objective is to understand how well SNN-based SDMs perform in comparison to conventional SDMs. Towards this, we have simulated both conventional and SNN-based SDM and CMM on Nengo. We observe that SNN-based models perform similarly as the conventional ones. In order to evaluate the performance of different SNNs, we repeated the experiment using Adaptive-LIF, Spiking Rectified Linear Unit, and Izhikevich models and obtained similar results. We conclude that it is indeed feasible to develop some types of associative memories using spiking neurons whose memory capacity and other features are similar to the performance without SNNs. Finally we have implemented an application where MNIST images, encoded with N-of-M codes, are associated with their labels and stored in the SNN-based SDM.

## I. INTRODUCTION

Sparse Distributed Memory (SDM) is a form of associative memory where data is encoded and stored using sparse representations. First introduced by Kanerva in 1998[1], it was subsequently implemented by Furber et. al. using N-of-M codes [2] as well as using Rank-order Codes [3]. The motivation of their work was to test the plausibility of implementing associative memories using large scale low-power hardware, which was also amenable to a spiking neural implementation and good memory. To achieve low-power, the environmental data needed to be encoded and stored in a sparse format. For this they used the N-of-M codes from Verhoeff [4], as well as the biologically inspired Rank-order codes, proposed by van Rullen and Thorpe [5]. Implementations of SNN-based SDMs have additionally been studied by Bose [6] and Sharp [7], although these did not focus on the memory capacity of such memories.

The objective of our work in this paper is to understand how spiking neurons can be used to implement SDMs, and what is the memory capacity achieved in comparison to the conventional SDMs. To that end, we have implemented N-of-M Correlation Matrix memories (CMMs) and SDMs with varying memory sizes and also implemented corresponding spiking SDMs. We measured the memory capacities of the spiking SDMs and found them to be consistent with the N-of-M SDMs. Furthermore, we perform a comparative analysis of the different kinds of SNNs used to build SDMs - such as LIF based SNNs, Adaptive-LIF SNNs, Izhikevich SNNs and Spiking Rectified Linear Unit based SNNs.

The paper is structured as follows: In Section 2, a review of the existing literature on associative memories and sparse distributed memories is provided. Section 3 provides a brief methodology of building the sparse distributed memory using spiking neural networks implemented on the Nengo framework. Section 4 covers the implementation details including the numerical values of all the constants, the spiking rates and the scaling factor used in our experiments. Section 5 presents the results of our experiments. Section 6 gives the details of an application involving storing MNIST images in the SDM. Section 7 concludes the paper.

## II. BACKGROUND

In this section we look at some background concepts and a short analysis of existing literature about the same.

### A. Correlation Matrix Memory (CMM)

Correlation Matrix Memory [8], first proposed by Kohonen in 1972, is a type of hetero associative memory - a neural network that is capable of storing associations between different patterns. All information about the correlation of the patterns is stored in the weight matrix of the associative memory.

Let there be $q$ patterns to be stored in the associative memory. Let $Y_k$ denote the $k^{th}$ pattern stored in the associative memory, and let $Y'_k$ denote the pattern recalled using the input pattern $X_k$ and the weight matrix of the associative memory $W_k$. Let the length of all output patterns be $n$, and length of all input patterns be $m$. Thus, all output patterns belong to $\{0,1\}^n$ and all input patterns belong to $\{0,1\}^m$. The associative memory can then be represented by (1):

$$y_k' = W_k * x_k \qquad (1)$$

where the weight matrix $W_k$ is represented as (2) after k X, Y pairs have been stored.

$$W_k = Y_k \cdot X_k^T \qquad (2)$$

The weight matrix of the CMM thus stores the sum of correlations between patterns. The CMM forms the pivotal data memory component of the SDM, which is described in the following section.

A visual representation of the CMM is shown in fig. 1.

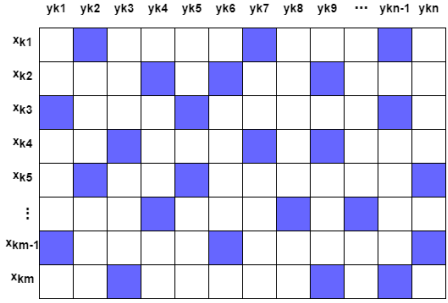

Fig. 1. A visual representation of a binary CMM. The coloured blocks indicate a correlation, and the blank squares imply no correlation.

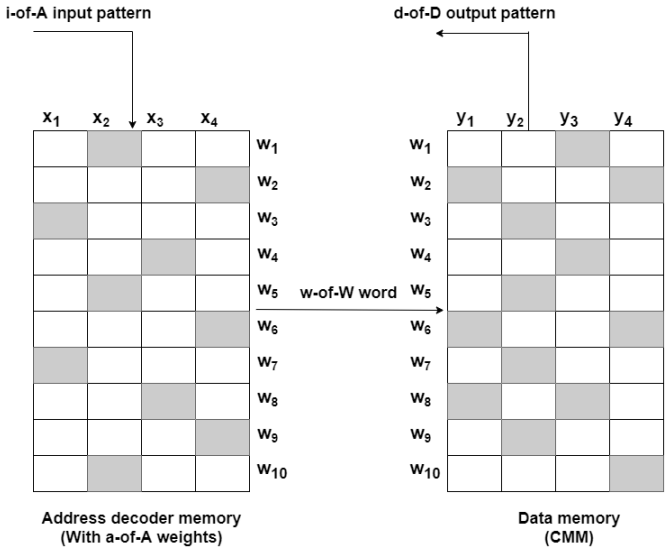

Fig. 2. A visual representation of a binary SDM. The first block represents the address decoder and the second a CMM.

### B. Sparse Distributed Memory (SDM)

SDMs were first proposed by Kanerva [1], as a sparse memory that could store distributed representations. The original SDM architecture proposed by Kanerva used binary vectors and had counters for memory locations. It had the advantages of robustness and noise tolerance. However the original SDM also had disadvantages such as increased memory utilization due to counters and processing time. Recently, Mendes et al [9] tried different encoding methods including gray code, over the original binary vectors proposed by Kanerva, and achieved better results for non-random data. Furber et. al [2] replaced counters with unipolar binary weights and obtained good memory capacity results, along with decreased processing time of the SDMs. In addition, they used CMMs, described in the previous subsection, as the data memory. Their encoding scheme was N-of-M coding, where exactly N out of total M bits are 1s and the rest are 0s.

The architecture of the SDM from Furber [2] is shown in fig. 2. It consists of a fixed weights address decoder layer, followed by a data memory layer, which is a CMM. Both the layers have binary weights and follow N-of-M codes.

The model has a read phase and write phase. During the write phase, two pieces of data (called 'address' and 'data'), both also N-of-M codes, are written to the memory together. The 'address' pattern is presented to the address decoder layer, which casts it into higher dimensional space, since the number of neurons W at the output is much higher than the number of neurons A at the input. The output of the address decoder is presented to the data memory, which is a CMM, along with the 'data' pattern. The CMM sets the weights of the data memory as described previously, thus associating the 'address' with the 'data'.

During the read phase, the 'address' is presented to the address decoder as earlier, and the address decoder output is presented to the data memory. This time, the weights of the data memory are not changed, rather each neuron in the data memory sums up the inputs as per the previously set connections. The top n-of-M neurons at the output of the data memory is said to constitute the output pattern. The SDM can thus store and retrieve a number of address-data pairs. It can sometimes retrieve the correct data pattern even if some noise is present in the address pattern.

In this paper, we have used N-of-M SDMs as the memory architecture, implementing them using spiking neural networks (SNNs) on the Nengo simulator [10].

### C. Leaky Integrate and Fire (LIF) Neurons

Leaky Integrate and Fire (LIF) neurons [11] is one of the most commonly used spiking neuron models, in part because of their simplicity combined with biological plausibility.

Derived from the biologically plausible Hodgkin-Huxley model [12], the LIF model strikes a balance between faster computation and biological plausibility. The LIF model represents a neuron as an R-C circuit, where $C_m$ is the membrane capacitance per unit area and $1/R_m$ represents the membrane conductivity. The model is mathematically represented in (3):

$$C_m * dV_m(t)/dt = I(t) - V_m(t)/R_m \quad (3)$$

Where $I(t)$ is the input current and $V_m(t)$ is the post-synaptic cell membrane voltage that evolves with time. When $V_m$ exceeds a threshold voltage $V_{thr}$, the neuron fires, which is called a spike event. The cell membrane voltage suddenly spikes to a high value ~30 mV, followed a sharp decrease, and then back to its resting $V_{rest}$.

In our work, we have used the LIF spiking neuron model, made available on the Nengo Simulator [11] to build the spiking SDM. We have used two variants of the LIF neurons available on Nengo, viz. the Adaptive LIF [13] and Rectified Linear LIF, to compare the memory capacity of the spiking SDM when these variants are used. In addition, we have also used another popular model of spiking neuron, Izhikevich model [14] for our comparative study. For the model, we have used the parameters corresponding to tonic spiking, since that is closest to the LIF model.

In the following subsection, we consider a couple of learning rules used for spiking neurons, which are also available in the Nengo simulator.

*D. Learning rules*

Learning rules in a network of spiking neural networks indicate how synaptic strengths in the connections between the neurons change over time. Furber [3] among others indicated how Hebbian style learning rules might be used to implement an SDM architecture using SNNs. Hebbian learning implies that the connection weights between two neurons are strengthened if they fire together, and weakened if not.

In our work, we have used two learning rules that are Hebbian and which are available on the Nengo simulator: The Bienenstock-Cooper-Munroe (BCM) learning rule [15] and the Oja learning rule [16].

**BCM Learning rule**: BCM [15] is a type of Hebbian learning rule which modifies the synaptic weights as a function of the presynaptic activity and the difference between the postsynaptic activity and the average postsynaptic activity. The driving force behind the working of the BCM learning rule is that if certain cells' activities are often driven above their average value, then those cells must be playing an important part in the neural network, and hence their synaptic connections should be strengthened as well.

**Oja learning rule**: The Oja learning rule [16] augments the Hebbian learning rule to account for a "forgetting" term which is a function of the weight of the connection and the postsynaptic activity. This "forgetting" term balances the growth of the weights, which could theoretically increase significantly in a simple Hebbian rule. This also makes the Oja rule slightly more biological plausible than the Hebbian rule.

In the following subsection, we briefly introduce the Nengo simulator and NengoDL.

*E. Nengo simulator and NengoDL*

Nengo [10] is a Python based simulator for spiking neurons. As a simulator, it has a few advantages: It offers flexibility in writing code to simulate spiking neurons, it has graphical and scripting interfaces, it works with multiple learning rules and spiking neural models, it is getting wider acceptance in the machine learning and computational neuroscience community. All of these inspired us to use Nengo for simulating our model.

NengoDL [17] is a simulator for SNNs that combines Nengo and Deep Learning. The input to NengoDL is a Nengo network. Using NengoDL, a user can simulate the SNN using a deep learning model such as Tensorflow, thus they can get the best of both worlds.

To switch from Nengo to NengoDL, only the simulator needs to be changed, while the model is same as that built for the Nengo simulator. NengoDL has a faster simulation speed on both CPUs and GPUs. Additionally, NengoDL allows batch processing, which significantly reduces the computation time.

In the following section, we explain our methodology for implementing CMMs and SDMs using spiking neurons and Nengo.

### III. METHODOLOGY

In this section, we describe the methodology for implementation of SNN based SDMs and CMMs.

Since we have based our work on previous work by Furber et al [2, 3] we have used a similar architecture of the SDM as used by them. We have described this architecture in our previous section. In this section we will focus on aspects related to the spiking (SNN) implementation.

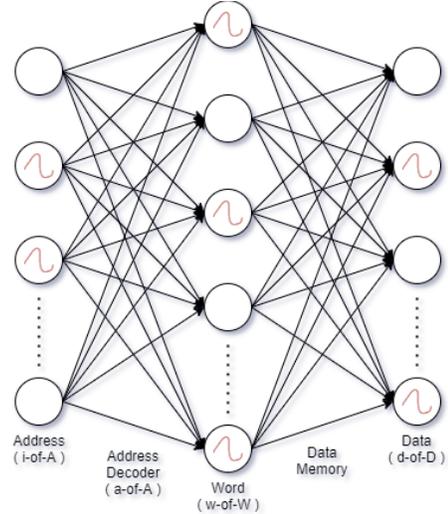

Fig. 3. A visual representation of a spiking SDM.

*A. Implementing a CMM with spiking neurons*

We have described the basic architecture of the CMM in a previous section. Here we assume all weights are binary, and so are the inputs and outputs.

The CMM consists of an input layer comprising *A* spiking neurons connected to an output layer of *D* spiking neurons in an all-to-all fashion. Similar to the SDM, the CMM also has a read phase and a write phase. All the inputs and outputs follow N-of-M coding. In spike terms, such a code can be approximated by stating that within a certain time window, N of the total of M inputs to a neuron fire. Similarly, only N neurons out of a layer of M neurons fire their output spikes within a certain time window. This can be implemented by using a inhibitory inter-neuron with threshold N, that fires a spike to cut off the neurons when exactly N inputs have fired. Such a mechanism has been discussed previously by Furber [3], Bose [6] and Sharp [7].

In our implementation, we have approximated the effect of the spikes by considering the rise in current of a neuron as an effect of an input spike. Hence, during the write phase, the inputs to the input and output layer are given as currents arising from the spikes.

Initially, the weights are initialized to zero. Each neuron is given a current of 1 unit or no current, depending on whether the input element is 1 or 0. This produces a spike train in the neurons that are provided a current, and the weights are modified using a Hebbian learning rule as described previously. As mentioned, in our implementation we have used the BCM [15] and Oja [16] learning rules.

Next, the weights of the neurons are multiplied by an amplification factor, denoted by $f_{BCM}$ and $f_{Oja}$ for the weights obtained by BCM and Oja learning rules respectively.

During the reading phase, we use the amplified weights, and provide a current only to the input layer, using the same vector as previously used in the writing phase. The spikes produced by the input layer correspond to a current of a certain value, 1000 in one case. This current gets multiplied with the weights and is supplied to the output layer neurons, which then generate spikes based on the amount of current provided, in accordance with the response curve of the neuron. From the $D$ neurons of the output layer, we select d neurons that fire first, corresponding to an N-of-M code ($d$-of-$D$ in this case). These positions are then set to 1, while the remaining positions are set to 0, thereby producing a binary output vector. Thus the output is a $d$-of-$D$ code.

### B. Implementing an SDM with spiking neurons

We have described the architecture of the binary SDM earlier in the previous section. There are two layers, address decoder and data memory, the first one of which has fixed weights and the second one is a CMM.

For the spiking SDM model, we replace each of the two memory structures, viz. address decoder and data memory with two layers of spiking neurons, connected in an all-to-all fashion.

**Address Decoder layer in the Spiking SDM**: The 'address decoder' consists of an input layer comprising $A$ neurons connected to an output layer of $W$ neurons. The weights are set in a similar manner as those for address decoder in the non-spiking SDM model and then multiplied by a factor $f_1$. The inputs to the input layer are given as currents arising from the spikes. Each neuron is given a current of 1 unit or no current, depending on whether the input element is 1 or 0. This simulates an input spike firing within a specific time window.

Only those neurons which are given a stimulus fire, and the remaining neurons do not fire. The spikes from the neurons that fire are multiplied with the address decoder weights, and summed up, to be fed as input current to the output layer. The output layer neurons that receive the highest input current have the highest firing rate. As expected, the neurons with the highest firing rate fire the earliest. The spike timings of the output layer are noted. The outputs of the neurons that fired the first w spikes are set to 1, while the outputs of the remaining neurons are set to 0. Thus, the output of the address decoder is a $w$-of-$W$ code, which serves as the input to the data memory layer.

**Data Memory layer in the Spiking SDM**: As mentioned earlier, the data memory in the SDM is a CMM. So it has a functionality similar to the CMM spiking implementation. It consists of an input layer comprising W neurons connected to an output layer of D neurons. The weights are initialized to zero, and are learnt using a Hebbian learning rule, as is done for spiking CMM. They are then multiplied by an amplification factor $f_2$. As mentioned earlier, we have worked with BCM and Oja learning rules. The read cycle of the data memory layer proceeds as in the CMM. As described in the spiking CMM, the output, during the write cycle is a $d$-of-$D$ code, representing d neurons that fire first out of a total of D neurons.

For the purpose of measuring how many address data pairs are stored and retrieved correctly, we use exact match: if the d-of-D code at the output during read phase exactly matches the code at the input, we count it as correct, else we count is as an error. In future, we will look at alternative measures such as Hamming distance.

### C. Memory capacity

The memory capacity for the SDM is computed as the point at which the memory starts making mistakes while recalling a previously stored address-data pattern. In order to determine the memory capacity, we vary the number of address data pairs stored in the memory and plot the number of pairs recalled correctly against the number of pairs that were written to the memory.

## IV. IMPLEMENTATION DETAILS

In this section, we mention a few items specific to our implementation of the SNN based SDM.

### A. Simulation platform

The Spiking CMM and SDM were simulated using the Nengo library [10] on Google Colaboratory [18]. During both the read and write phases of the SDM, we ran our Spiking SDM simulations for 150 milliseconds, in order to accurately obtain the spikes from the neurons over time. However, we have used only the first spike from each neuron to determine to first w or first d spikes for address decoder and data memory respectively.

### B. Numerical values of the parameters

For our experiments, we have chosen 11-of-256 codes for both the address and data vectors, i.e. i= 11 and A=D= 256.

We have chosen the value of a to be 20 for all experiments. The number of neurons in the hidden layer (W) has been varied to test the scalability, and we have taken W to be equal to 256, 512, and 1024 for our experiments. However, the value of w is kept constant at 16 for all experiments.

We have used the default parameters for different spiking models available with Nengo. For example, for Izhikevich model, we used *nengo.Izhikevich()* which uses Tonic spiking parameters by default [19]. We similarly used default Nengo parameters for LIF, Adaptive LIF and Spiking RELU models [20].

### C. Spiking rates

We aimed to modulate the hyperparameters of the SNN in such a way that the spiking rate should be neither too high nor too low. In case the spiking rate is too high, the neuron will spike at approximately the same time for close values of inputs, and it would be impossible to decode the output to discern which d neurons were the first ones to spike. A very low spiking rate, on the other hand, would produce the first spike beyond 150 milliseconds. Since we only consider the spikes fired in the first 150 milliseconds, most neurons would display no spikes, leading to a wrong output after decoding. We choose a maximum spiking rate of 100 spikes per second for the LIF and Adaptive-LIF neurons. For the Spiking ReLU neuron, we set the maximum spiking rate to be 50 spikes per second. Finally, for the Izhikevich neuron, we select a maximum spiking rate of 10 spikes per second. These spiking rates have been adjusted empirically, using the response curves of the various neurons.

*D. Scaling factor*

The current produced by a single spike fired by the neurons of the input layer is numerically equal to 1000. These are added and can theoretically go up to w*1000 at the address decoder output and d*1000 at the data memory output. The weights of the address decoder of the Spiking SDM model must, therefore, be scaled down to constrain the inputs at the respective output layers, preferably to a range of 0-10. Additionally, the weights produced using the learning rules are in the order of $10^{-6}$ and $10^{-5}$ for BCM and Oja rules respectively, which need to be scaled up. For the CMM, we use the scaling factors as $f_{BCM} = 100$ and $f_{Oja} = 16$ for the weights obtained using BCM and Oja learning rules respectively. For the SDM, we use a scaling factor of $f_1 = 1/250$ for the address memory weights and $f_2 = 60$ for the data memory weights obtained using the BCM learning rule. These values were determined empirically.

In the following section, we describe the results we obtained in our experiments.

## V. RESULTS

In this section, we describe the results of memory capacity of the SDM and CMM using spiking neurons, that we obtained in our experiments.

*A. Memory capacity of CMM using the BCM and Oja learning rules*

The memory capacity of the CMM using BCM and Oja learning rules is plotted in figures 4 and 5 respectively.

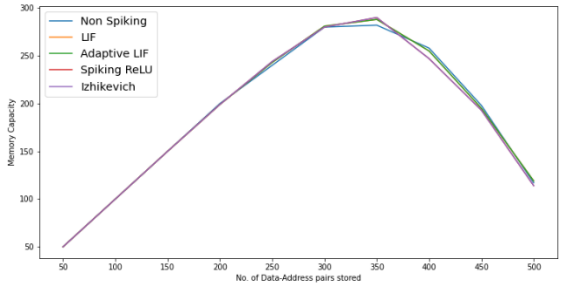

Fig. 4. Plot of the memory capacity of the spiking neural CMM using the BCM learning rule on Nengo, with different spiking neural models.

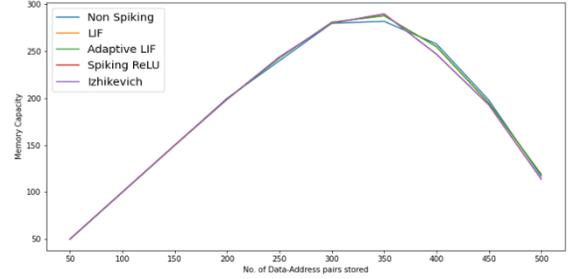

Fig. 5. Plot of the memory capacity of the spiking neural CMM using the Oja learning rule on Nengo, with different spiking neural models.

We can make the following observations from the CMM memory capacity plots:

- Different SNN models, including LIF, Adaptive LIF, ReLU and Izhikevich, give similar results in terms of memory capacity.

- The SNN models and the non-spiking model (i.e. conventional CMM without spiking neurons) give similar results.

- BCM and Oja learning rules give similar results in terms of memory capacity.

- Until we have written around 300 address-data pairs, the recall is perfect. Between 300 to 350, the memory capacity starts to plateau out, and writing additional address-data pairs does not give any better recall. Beyond 350, the memory capacity falls, although the fall is gradual rather than catastrophic.

*B. Memory capacity of SDM using the BCM learning rule, varying the memory size*

For the SDM, we used only the BCM learning rule. The memory capacity for varying memory sizes (256, 512 and 1024) is plotted in figures 6 to 8.

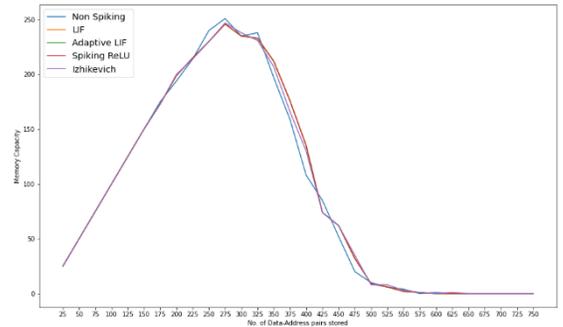

Fig. 6. Plot of the memory capacity of the spiking neural SDM using the BCM learning rule on Nengo, for a address decoder layer size of 256 (encoded as 16-of-256 vectors), with different spiking neural models.

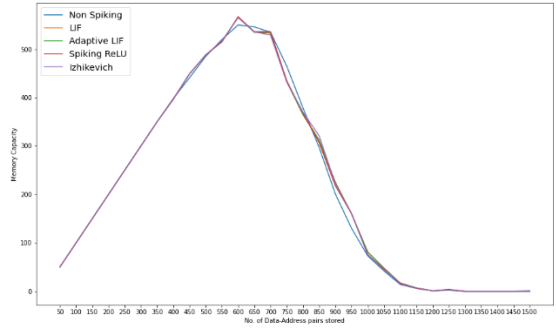

Fig. 7. Plot of the memory capacity of the spiking neural SDM using the BCM learning rule on Nengo, for an address decoder size of 512 (encoded as 16-of-512 vectors), with different spiking neural models.

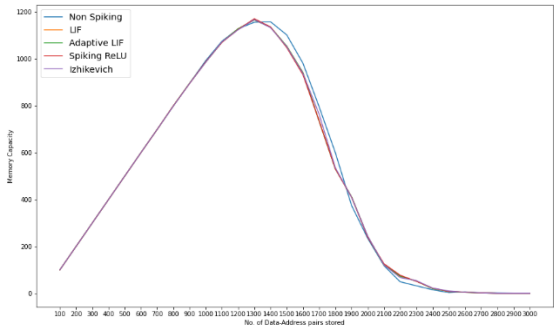

Fig. 8. Plot of the memory capacity of the spiking neural SDM using the BCM learning rule on Nengo, for a address decoder size of 1024 (encoded as 16-of-1024 vectors), with different spiking neural models.

We can make the following observations from the SDM memory capacity figures 6-8:

- Different SNN models, including LIF, Adaptive LIF, ReLU, Izhikevich tonic spiking model, give similar results in terms of memory capacity.
- The SNN models and the non-spiking model (i.e. conventional CMM without spiking neurons) give similar results.
- For different varying memory sizes, the memory capacity plots follow a similar pattern. For smaller number of address-data pairs, the recall is perfect i.e. the graph grows linearly. At a certain stage, the memory capacity starts to plateau out, and writing additional address-data pairs does not give any better recall. Beyond that stage, the memory capacity falls, although the fall is gradual rather than catastrophic.
- The CMM and SDM memory capacity plots are broadly similar, although the size at which the memory capacity is reached is higher for the SDM because of the addition of the extra address decoder layer compared to the CMM.
- The memory capacity plots for different memory sizes shows similar trends. From this, we conclude that the SDM and CMM memories are scalable.

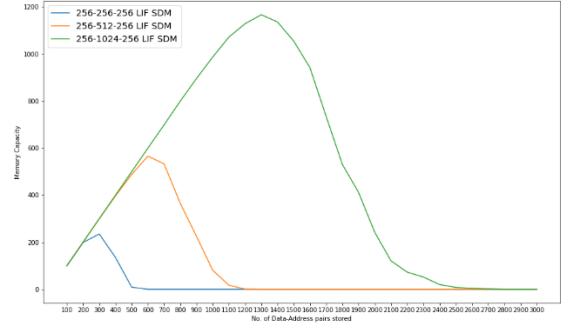

Fig. 9. Plot of the memory capacity of the spiking neural SDM using an LIF neuron and the BCM learning rule on Nengo, for and varying the address decoder size from 256 to 1024.

Figure 9 shows a plot of the memory capacity with varying SDM sizes of the address decoder from 256 to 1024, plotted in a single graph to study how the memory behaves when the memory size is increased and other parameters such as SNN model are kept constant. As we can see, the shape is similar for all the increasing memory sizes. In each case, the memory capacity reaches a peak and then degrades slowly beyond that peak.

From this, we can conclude the SNN based SDM can store and retrieve memories successfully and is also scalable.

In the following section, we look at a simple application using the SNN based SDM.

## VI. APPLICATION: STORING MNIST IMAGES

In this section, we give details of a simple application using storage of images in the SNN based SDM. Our motivation was to display the capability of the SNN based SDM in a practical application.

For the implementation, we took images of handwritten numbers from the MNIST dataset [21], along with the labels of the numbers corresponding to those images. We fed the image to an N-of-M encoder neural network, that converted each 28*28 MNIST image to a 22-of-256 code, to feed to the SDM.

The encoder weights were fixed and the output was an N-of-M code corresponding to the MNIST image. This served as the 'address' in the SDM, while the 'data' was the label corresponding to each MNIST image, converted to a 1-of-10 code. The write and read phases were implemented as before using the SNN-based SDM and the Nengo simulator.

The flow of the write and read phases using the SNN based SDM is shown in figure 10.

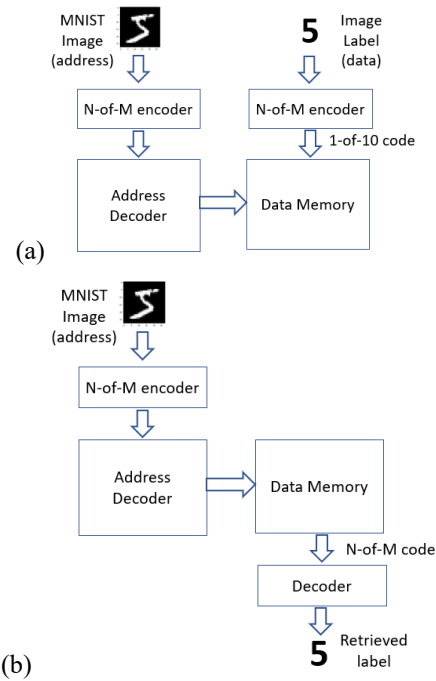

Fig. 10. Illustration of the application for storing MNIST images and retrieving their label using SNN based SDM during (a) write phase and (b) read phase. .

With the SNN-based SDM, we were able to store and retrieve the associations of the labels with MNIST images.

Figure 11 shows the plot of the memory capacity, for an SDM with input of size 256 and middle layer of size 512. The memory capacity grows linearly (i.e. perfect recall is achieved) until around 400 address-data pairs, after which it starts degrading slowly.

However, a peak could not be discerned in the memory capacity graph. We simulated the memory until we reached 3000 address data pairs.

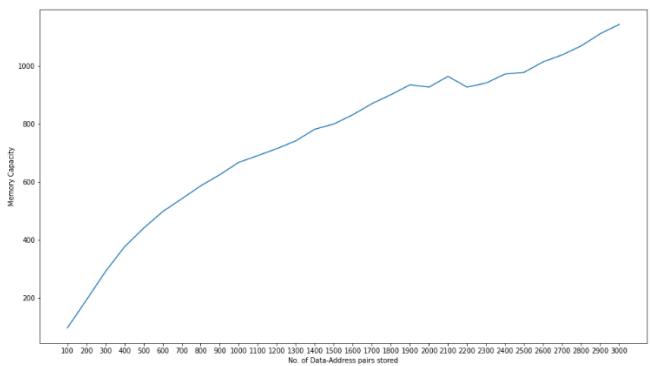

Fig. 11. Plot of the memory capacity for the MNIST images application

One reason why we could not see a clear peak in the memory capacity plot is because the data is only one of a fixed number of values (0 to 9) even though the number of addresses (MNIST images) is more, and hence the memory capacity is vastly increased as the memory learns to generalize from the address-data mapping.

## VII. CONCLUSION AND FUTURE WORK

We have presented implementations of the CMMs and SDMs using spiking neurons. We used the Nengo simulator for our implementation, and the biologically plausible BCM and Oja learning rules for training the SNNs.

We used the CMM as the data memory of the SDM model. The address memory of the SDM was built using randomly initialized unipolar binary weights. We observed that the spiking SDM gave a similar memory capacity as the non-spiking SDM model.

We then tested the model for various address decoder sizes varying from 256 to 1024, and found it to be scalable with similar performance. Furthermore, we experimented with different spiking neurons viz. LIF, Adaptive-LIF, Izhikevich and Spiking ReLU, and found that the choice of neuron model did not have a significant effect on the memory capacity. We also implemented a simple application using MNIST images stored in the SDM, and found that the images were retrieved successfully.

In future, we would like to test the memory for some real life applications such as storing encoded images or text. For this purpose, the images or text could be encoded using spiking neural network into a binary vector and be stored using the SDM model using significantly less memory than conventional systems. This might be specially suitable for Internet of Things (IoT) related applications, where the properties of SNN based SDMs such as low power and good memory capacity could prove to be useful.

Overall, we believe that our preliminary study on spiking neural network based SDMs will contribute to the current research in biologically-inspired associative memory models.